\documentclass[twoside,leqno,twocolumn]{article}  
\usepackage{ltexpprt} 

\usepackage{times,epsfig,graphicx}
\usepackage{algorithm,algorithmic}

\usepackage[latin1]{inputenc}
\usepackage{fancyhdr}
\usepackage{verbatim}
\usepackage{tabularx}
\usepackage{hyperref}
\usepackage{xspace}
\usepackage{bm}

\usepackage{amssymb,amsmath}
\makeatletter

\DeclareRobustCommand\onedot{\futurelet\@let@token\@onedot}
\def\@onedot{\ifx\@let@token.\else.\null\fi\xspace}

\def\etc{\emph{etc}\onedot}  
 
\def\etal{\emph{et al}\onedot}

\newcommand{\rar}{\rightarrow}

\def\eq#1{(\ref{#1})}
\def\be{\begin{equation}}
\def\ee{\end{equation}}
\def\bea{\begin{eqnarray}}
\def\eea{\end{eqnarray}}
\def\ben{\begin{eqnarray*}}
\def\een{\end{eqnarray*}}

\def\bi{\begin{itemize}}
\def\ei{\end{itemize}}

\newcommand{\bt}[1]{\begin{tabular}{#1}}
\newcommand{\et}{\end{tabular}}
\newcommand{\ba}[1]{\begin{array}{#1}}
\newcommand{\ea}{\end{array}}
\newcommand{\ul}[1]{\underline{#1}}

\def\Loss{\mathcal L}

\DeclareMathOperator*{\argmin}{argmin}

\def\Re{{\rm I\!R}}                            
\def\<{\langle}
\def\>{\rangle}

\def\eye{{\bf I}}
\newcommand{\bfone}{{\bf 1}}

\newcommand{\bfx}{{\bf x}}

\newcommand{\bfZ}{{\bf Z}}
\newcommand{\bfX}{{\bf X}}
\newcommand{\tbfX}{\tilde{\bf X}}
\newcommand{\bfY}{{\bf Y}}

\newcommand{\bfW}{{\bf W}}
\newcommand{\bfP}{{\bf P}}
\newcommand{\bfQ}{{\bf Q}}
\newcommand{\bfS}{{\bf S}}

\newcommand{\bfA}{{\bf A}}
\newcommand{\bfB}{{\bf B}}
\newcommand{\bfC}{{\bf C}}

\newcommand{\bfK}{{\bf K}}
\newcommand{\bfM}{{\bf M}}
\newcommand{\bfN}{{\bf N}}

\newcommand{\calD}{{\cal D}}
\newcommand{\calT}{{\cal T}}

\newcommand{\E}{\mathbb{E}}

\newcommand{\Exp}{\mathbb{E}}

\begin{document}

\title{An Extended Framework for Marginalized Domain Adaptation}

\author{Gabriela Csurka, Boris Chidlovski,  St\'{e}phane Clinchant and Sophia Michel \\
Xerox Research Center Europe (XRCE), \\
6 chemin Maupertuis, 38240 Meylan, France, \\
\url{Firstname.Lastname@xrce.xerox.com}}
\date{}

\maketitle


\begin{abstract} \small\baselineskip=9pt  
  {\em We propose an extended framework for marginalized domain adaptation,
  aimed at addressing unsupervised, supervised and semi-supervised
  scenarios. We argue that the denoising principle should be extended
  to explicitly promote domain-invariant features as well as help the
  classification task. Therefore we propose to jointly learn  the data
  auto-encoders and the target classifiers. First, in order to make the
  denoised features domain-invariant, we propose a domain
  regularization that may be either a domain prediction loss or
  a maximum mean discrepancy between the source and target data. The
  noise marginalization in this case is reduced to solving the linear
  matrix system $AX=B$ which has a closed-form solution. Second, in
  order to help the classification, we include a class
  regularization term. Adding this component reduces the learning problem
  to solving a Sylvester linear matrix equation $AX+BX=C$, for which
  an efficient iterative procedure exists as well. We did an extensive
  study to assess how these regularization terms improve the
  baseline performance in the three domain adaptation scenarios.
  We present experimental results
  on two image and one text benchmark datasets, conventionally used
  for validating domain adaptation methods. We report our findings and
  comparison with state-of-the-art methods. }
\end{abstract}

\section{Introduction}

While huge volumes of unlabeled data are generated and made available
in many domains, the cost of acquiring data labels remains high.
Domain Adaptation (DA) problems arise each time we need to leverage
labeled data in one or more related {\it source} domains, to learn a
classifier for unseen or unlabeled data in a {\it target} domain.  The
domains are assumed to be related, but not identical and this
{\it domain shift} occurs in multiple real-world applications,
such as named entity recognition or opinion extraction across
different text corpora, \etc.

In this paper, we build on the domain adaptation work based on noise
marginalization~\cite{ChenICML12Marginalized}. In deep learning, a
denoising autoencoder (DA) learns a robust feature representation from
training examples. In the case of domain adaptation, it takes 
unlabeled instances of both source and target data and learns a new
feature representation by reconstructing the original features from
their noised counterparts. A {\it marginalized denoising autoencoder}
(MDA) marginalizes the noise at training time and thus does not
require an optimization procedure using explicit data corruptions to
learn the model parameters but computes the model in closed form.
This makes MDAs scalable and computationally faster than the regular
denoising autoencoders. The principle of noise marginalization has
been successfully extended to learning with corrupted
features~\cite{MaatenICML13Learning}, link prediction and multi-label
learning~\cite{ChenAAAI15Marginalized}, relational
learning~\cite{ChenSICDM14Marginalized}, collaborative
filtering~\cite{LiCIKM15Deep} and heterogeneous cross-domain
learning~\cite{li16Learning,ZhouAAAI14Hybrid}.

In this paper we extend the previous efforts and propose a larger
framework for the marginalized domain adaptation. The {\it
marginalized domain adaptation} refers to a denoising of source and
target instances that explicitly makes their features {\it domain
invariant} and eases the target prediction. 
We propose two extensions to the MDA. The first extension is a domain 
regularization, aimed at generating domain invariant
features. Two families of such regularization are considered; one is
based on the {\it domain prediction principle}, inspired by the
adversarial learning of neural networks~\cite{GaninX15Domainadversarial}; 
the second uses the maximum mean discrepancy (MMD) measure~\cite{HuangNIPS07Correcting}.

The second extension to the MDA is a class regularization; it allows to generate a 
classifier for target instances which can be learned jointly with the domain invariant 
representation.

Our framework works in {\it supervised,
unsupervised and semi-supervised} settings, where the source data is
completed with a few labeled target data, massive unlabeled target
data or both, respectively. In all cases, the noise marginalization is
maintained, thus ensuring the scalability and computational
efficiency.
We show how to jointly optimize the data
denoising and the domain regularization, and how to marginalize the
noise, which guarantees the closed-form solution and thus the
computational efficiency.  When the joint optimization is extended to
the target prediction, the solution does not have a closed form, but
is the solution of a {\rm Sylvester linear matrix equation} 
$AX+XB=C$, for which efficient iterative methods can be used.


The remainder of the paper is organized as follows.  In
Section~\ref{sec:soa} we revise the prior art. Section~\ref{sec:da}
presents the components of the marginalized domain adaptation,
including instance denoising, domain and class regularizations and target
classifier learning. The joint loss minimization is detailed in
Section~\ref{sec:joint}. In Section~\ref{sec:eval} we describe two
image and one text datasets we used, the experimental settings. We 
report the evaluation results which are grouped and
analyzed by the three settings, namely, unsupervised, supervised and
semi-supervised ones. Section~\ref{sec:conclusion} discusses the open
questions and concludes the paper.

\section{State of the art}
\label{sec:soa}

Domain adaptation for text data has been studied for more than a
decade, with applications in statistical machine translation, opinion
mining, and document ranking~\cite{DaumeJAIR06Domain,ZhouSIGKDD14Unifying}.
Most effective techniques include feature
replication~\cite{DaumeX09Frustratingly}, pivot
features~\cite{BlitzerEMNLP06Domain} and finding
topic models that are shared between source and target
collections~\cite{ChenICML14Topic}. Domain
adaptation has equally received a lot of attention in computer vision\footnote{For a recent comprehensive  survey see \url{http://arxiv.org/abs/1702.05374}}.
A considerable effort to systematize different shallow domain adaptation and
transfer learning techniques has been undertaken
in~\cite{GopalanFTCGV15Domain,PanTKDE10Survey,CsurkaX17Comprehensive}. These
studies distinguished three main categories of domain adaptation
methods.  The first category aims at correcting sampling bias
~\cite{XuNIPS10Multisource}.
The second category is in line
with multi-task learning where a common predictor is learned for all
domains, which makes it robust to domain
shift~\cite{ChenNIPS11Cotraining}.  The third family seeks to find a
common representation for both source and target examples so that the
classification task becomes easier~\cite{PanTNN11Domain}.
Finally, an important research direction deals with the theory of domain 
adaptation, namely when adaptation can be effective and guaranteed with 
generalization bounds~\cite{BendavidNIPS07Analysis}.

More recently, deep learning has been proposed as a generic solution
to domain adaptation and transfer learning
problems \cite{ChopraWREPL13Dlid,GlorotICML11Domain,LongICML15Learning}.
One successful method which aims to find common features between
source and target collection relies on {\it denoising autoencoders}.
In deep learning, a denoising autoencoder is a one-layer neural
network trained to reconstruct input data from partial random
corruption~\cite{VincentICML08Extracting}.  The denoisers can be
stacked into multi-layered architectures where the weights are
fine-tuned with costly back-propagation.  Alternatively, outputs of
intermediate layers can be used as input features to other learning
algorithms. This learned feature representation was applied to
domain adaptation~\cite{GlorotICML11Domain}, where stacked denoising
autoencoders (SDA) achieved top performance in sentiment analysis
tasks. The main drawback of SDAs is the long training time, and Chen \etal~\cite{ChenICML12Marginalized} proposed a variant of SDA where the
random corruption is {\it marginalized out}.  This crucial step yields
a unique optimal solution which is computed in closed form and
eliminates therefore the need for back-propagation.  In addition, features
learned with this approach lead to a classification accuracy
comparable with SDAs, with a remarkable reduction of the training
time~\cite{ChenICML12Marginalized}. 
 
More recently, deep learning architectures have demonstrated their
ability to learn robust features and that good transfer performances
could be obtained by just fine-tuning the neural network on the target
task~\cite{ChopraWREPL13Dlid}.  While such solutions perform
relatively well on some tasks, the refinement may require a
significant amount of new labeled data.  More recent works proposed
better strategies than fine-tuning, by designing deep architecture for
the domain adaptation task. For example,
Ganin \etal~\cite{GaninX15Domainadversarial}
has shown that adding a domain prediction task while learning the deep
neural network leads to better domain-invariant feature
representation.  Long \etal~\cite{LongICML15Learning} proposed to add
a multi-layer adaptation regularizer, based on a multi-kernel maximum mean 
discrepancy (MMD). These approaches obtained a
significant performance gain which shows that transfer learning is not
completely solved by fine-tuning and that transfer tasks should be
addressed by appropriate deep learning representations.

\section{Domain adaptation by feature denoising}
\label{sec:da}

We define a domain $\calD$ as the composition of a feature space
${\cal X} \subset \Re^d$ and a label space $\cal Y$. A given task in
the domain $\cal D$ (classification, regression, ranking, etc.) is
defined by a function $h: \cal X \rar \cal Y$.  In the domain
adaptation setting, we assume working with a source domain $\calD^s$
represented by the feature matrix $\bfX^s$ and the corresponding labels $\bfY^s$, 
and a target domain $\calD^t$ with the features $\bfX^t$.

We distinguish among three scenarios of domain adaptation, depending
on what is available in the target domain:
\begin{itemize}
\item {\it Unsupervised (US)} setting, where all available target instances are 
unlabeled. In this case, $\bfX_l^t$ is empty and the labeled data, denoted 
by $\bfX_l$ contain only the labeled source examples, $\bfX_l=\bfX^s$.

\item {\it Supervised (SUP)} setting, where few
 labeled target instances $\bfX_l^t$ are available at training
 time. In this case, we have $\bfX_l=[\bfX^s,\bfX_l^t]$.

\item {\it Semi-supervised (SS)} setting, where 
massively unlabeled ($\bfX_u^t$) target data are available together
with few labeled ($\bfX_l^t$) data at the training time.
\end{itemize}

In what follows we propose a framework to address all three scenarios
in one uniform way. It aims at finding such a transformation of source
and target data 
that minimizes the following loss function:
\begin{equation}
\Loss= \Loss_1 + \lambda \Loss_2 + \gamma \Loss_3,
\label{eq:3terms}
\end{equation}
where 
\begin{itemize}
\item $\Loss_1$ is the data denoising loss on all data $\bfX= [\bfX^s,\bfX^t]$, 
\item $\Loss_2$ is the cross-domain classification loss on labeled data 
$\bfX_l=[\bfX^s,\bfX_l^t]$ with labels $\bfY_l=[\bfY^s,\bfY_l^t]$, 
\item $\Loss_3$ is the domain regularization loss on source and target data $\bfX$.
\end{itemize} 
Parameters $\lambda$ and $\gamma$
capture the trade-off between the three terms. All losses and parameters
are described in the following subsections. Intuitively,
minimizing the total loss~\eq{eq:3terms} can help exploring the
implicit dependencies between the data denoising, the domain
regularization and the cross-domain classification.

In this paper we study the case when all three terms
in \eq{eq:3terms} belong to the class of squared loss
functions\footnote{Other loss functions such as exponential,
logistic, hinge losses are interesting to explore as well, but they are 
beyond of the scope of this paper.
}. More precisely:
\begin{itemize}
\item $\Loss_1\equiv \Loss_1(\bfX, \bfW)$ is the {\it instance denoising loss} under the dropout law; 
we minimize the square loss $\|\bfX -\tilde \bfX \bfW\|^2$ between the
corrupted data $\tilde \bfX$ and the original data $\bfX$ denoised
with the linear transformation $\bfW$.  This term is the core element
of the marginalized denoising autoencoder (MDA)~\cite{ChenICML12Marginalized}.

\item $\Loss_2\equiv \Loss_2(\bfX_l,\bfY_l,\bfW,\bfZ_l)$ is the {\it class regularization loss}, 
aimed at learning a (multi-class) ridge classifier $\bfZ_l$ from the available
corrupted and denoised instances ${\tilde \bfX_l} \bfW$. The term is defined as $\|\bfY_l
-{\tilde \bfX_l} \bfW \bfZ_l\|^2$. It can be seen as a
generalization of the Marginalized Corrupted Features (MCF)
framework~\cite{MaatenICML13Learning} with a square loss (the MCF corresponds to 
the case when $\bfW=\eye_d$).

\item $\Loss_3 \equiv \Loss_3(\bfX^s,\bfX^t,\bfW)$ is the {\it domain regularization loss} that 
expresses the discrepancy between the source and target domains. We explore two
options for this term. One is based on the empirical maximum mean
discrepancy (MMD), taking into account the class labels when available; the 
other uses a pre-trained domain classifier to regularize the total loss.

\end{itemize}

We follow the marginalized framework for optimizing the loss on
corrupted data~\cite{ChenICML12Marginalized,MaatenICML13Learning}, and minimize 
the loss expectation $\Exp[\Loss]$. To simplify the reading, we
denote the expected loss values $\Exp[\Loss_i]$ also with $\Loss_i$.

By minimizing the marginalized expected loss~\eq{eq:3terms}, $\argmin_{\bfW,\bfZ_l} \Loss$, 
we obtain optimal solutions for the transformation matrix $\bfW$ and classifier $\bfZ_l$. 
This can be achieved in two different ways, namely:
\begin{itemize}
\item $\bfW$ and $\bfZ_l$ are learned {\em sequentially}. In this case we first 
set  $\lambda=0$ and learn $\bfW$ by minimizing $\Loss_1 + \gamma \Loss_3$. Then for the 
fixed $\bfW$ we learn $\bfZ_l$ from $\Loss_2$. 
Except the supervised MMD in $\Loss_3$, 
the learning of $\bfW$ remains unsupervised, including 
the supervised and semi-supervised settings. The target
labels in these cases are used at the second step, when the
classifier $\bfZ_l$ is learned.

\item $\bfW$ and $\bfZ_l$ are learned {\em jointly}.  In this case we iteratively 
optimize the joint loss with respect to $\bfW$ and $\bfZ_l$. To initialize the 
iterative process, we set $\bfW=\eye_d$ and minimize $\Loss$ to compute
$\bfZ_l$, then we fix $\bfZ_l$ and optimize $\Loss$ with respect to
$\bfW$, and so on. The process is repeated until convergence. In practice we
observed that the convergence is achieved after several iterations. 
\end{itemize}

In the following subsections we describe in details and discuss each of
the three loss terms, in Section~\ref{sec:joint} we address
their different combinations.

\subsection{Domain Instance Denoising}
\label{ssec:da}

The first term we consider is the loss used by the {\it Marginalized
Denoising Autoencoder} (MDA)~\cite{ChenICML12Marginalized}.
Its basic idea is to reconstruct the input data from a partial random
corruption~\cite{VincentICML08Extracting} with a marginalization that
yields optimal reconstruction weights in a closed form. The MDA loss
can be written as
\be
\label{eq:mda}
\Loss_1 \equiv \frac{1}{M} \sum_{m = 1}^M \| \bfX -\tbfX_m \bfW \|^2 + \omega \| \bfW \|^2,  
\ee
where $\tbfX_m \in \Re^N \times \Re^d$ is the $m$-th corrupted version
 of $\bfX$ by random feature dropout with a probability $p$ and
 $\omega \|\bf W \|^2$ is a regularization term. In order to avoid the
 explicit feature corruption and an iterative optimization,
 Chen \etal~\cite{ChenICML12Marginalized} showed that by
 considering the limit case $M\rightarrow \infty$, the weak law of
 large numbers allows to rewrite the loss $\Loss_1$ as its
 expectation and the optimal $\bfW$ can be written as (see  Appendix for details):
\be
\bfW=(\bfQ +\omega  \eye_d)^{-1} \bfP,
\label{eq:WMDA}
\ee
where $\bfP$ and $\bfQ$ depend only on the covariance matrix
$\bfX^{\top} \bfX$ and the noise level $p$. 

One main advantage of the MDA is that it requires no label 
and therefore can be applied in all three settings
{\it US, SUP} and {\it SS}. Note in the supervised case
$\bfX_l=[\bfX^s,\bfX_l^t]$ includes only few target examples to
learn $\bfW$.

\subsection{Learning with marginalized corrupted features}
\label{sec:mcf}

Inspired by the Marginalized Corrupted Features (MCF)
approach~\cite{MaatenICML13Learning}, we propose to marginalize the
following loss:
\begin{equation}
\Loss_2 \equiv \frac{1}{M}  \sum_{m=1}^M
\|\bfY_l - \tbfX_{lm} \bfW \bfZ_l\|^2  + \delta \|\bfZ_l\|^2,
\label{eq:mcf}
\end{equation}
where $\bfZ_l \in \Re^d \times \Re^C$ is a multi-class classifier
(each column corresponds to one of the $C$ classes),
$\bfY_l\in \Re^N \times \Re^C$ is a label matrix, where $y_{nc}=1$ if
$\bfx_n$ belongs to class $c=1,\ldots,C,$ and -1 otherwise, and
$\delta \|\bfZ_l\|^2$ is a regularization term. 
When $\bfW=\eye_d$, we obtain the MCF baseline where the classifier is
learned directly with the corrupted features. Moreover, when $p=0$, we obtain 
the ridge classifier learned with the original features. 

Given $\bfW$, the multi-class classifier $\bfZ_l$ can be computed in
closed form using the expected loss of~\eq{eq:mcf} (see derivations
in the Appendix):
\begin{equation}
\bfZ_l^*=  (1-p)(\bfW^\top \bfQ_l \bfW + \delta \eye_d)^{-1} \bfW^\top \bfX_l^\top \bfY_l.
\label{eq:Zl}
\end{equation}
The computation of $\bfZ_l^*$ requires the labeled data $\bfX_l$, that
contain the source ({\it US}) or possibly target data ({\it SUP, SS}).

\subsection{Reducing the discrepancy between domains}
\label{sec:mmd}

The domain regularization term $\Loss_3$ in~\eq{eq:3terms} is aimed at bringing
the target domain closer to the source domain, by minimizing the
discrepancy between the domains. In the following, we explore three options for
the term $\Loss_3$, namely (1) a classical empirical MMD using the
linear kernel, (2) its supervised version where the discrepancy is
minimized between the class means and (3) a domain classifier
$\bfZ_\calD$ trained on the uncorrupted data to distinguish between
the source and target data.

\begin{table*}[t]
\begin{center}
\caption{All models of the extended framework, with the corresponding notations, losses and solutions.}
\label{tab:losses}
\begin{tabular}{l | c | c } \hline
Method & Loss &  Closed-form solution for $\bfW$ \\ 

\hline
{\bf S1}   & $\Loss_1$
& $(\bfQ+\omega \eye_d)^{-1}\bfP$ \\

{\bf S1M}  & $\Loss_1 + \gamma \Loss_m$
& $(\bfQ+\omega \eye_d+ \gamma  \bfM)^{-1}\bfP$ \\

{\bf S1C}  & $\Loss_1 + \gamma \Loss_c$
& $(\bfQ+\omega \eye_d+ \gamma  \bfM_c)^{-1}\bfP$ \\

{\bf S1D}  & $\Loss_1 + \gamma \Loss_d$
& $\bfQ^{-1} (\bfP+\gamma (1-p)\bfX^\top \bfY_\calT \bfZ_\calD^\top)(\eye_d + \gamma \bfZ_\calD \bfZ_\calD^\top)^{-1}$ \\
\hline

\hline
&  & $\bfW$ as solution of $\bfA \bfW + \bfW \bfB =\bfC$  \\ 
\hline
{\bf J12}  & $\Loss_1 +\lambda \Loss_2$
& $\bfA=\bfA_{12}=\bfQ_l^{-1}(\bfQ+\omega \eye_d)$,
$\bfB=\bfB_{12}=\lambda \bfZ_l \bfZ_l^\top$, \\
&
&  $\bfC=\bfC_{12}=\bfQ_l^{-1} (\bfP + \lambda (1-p) \bfX_l^\top  \bfY_l \bfZ_l^\top)$ \\

\hline
{\bf J12M} & $\Loss_1 +\lambda \Loss_2 + \gamma \Loss_m$
& $\bfA =\bfQ_l^{-1}(\bfQ+\omega \eye_d+ \gamma \bfM),\bfB=\bfB_{12}$, $\bfC=\bfC_{12}$ \\
\hline
{\bf J12C} & $\Loss_1 +\lambda \Loss_2 + \gamma \Loss_c$
& $\bfA =\bfQ_l^{-1}(\bfQ+\omega \eye_d+ \gamma \bfM_c), \bfB=\bfB_{12}$, $\bfC=\bfC_{12}$ \\

\hline
{\bf J12D} & $\Loss_1 +\lambda \Loss_2 + \gamma \Loss_d$
& $\bfA=\bfQ_l^{-1}\bfQ$,
$\bfB=\bfA_{12} (\eye_d+ \gamma \bfZ_\calD \bfZ_\calD^\top)^{-1}$, \\
&
&  $\bfC=(\bfC_{12} + \bfQ_l^{-1} \gamma (1-p)\bfX^\top \bfY_\calT \bfZ_\calD^\top)(\eye_d + \gamma \bfZ_\calD \bfZ_\calD^\top)^{-1}$           \\

\hline
\end{tabular}
\end{center}
\end{table*}

\subsubsection{\bf Reducing the MMD between domains}
\label{ssec:mmd-us}


The minimization of maximum mean discrepancy (MMD)~\cite{Borgwardt06Integrating} between 
the source and target domains is the state of art approach widely used in the literature. It is 
often integrated in feature transformation learning~\cite{BaktashmotlaghICCV13Unsupervised,PanTNN11Domain} or
used as a regularizer for the cross-domain classifier
learning~\cite{DuanPAMI12Domain,LongICML15Learning,TzengX14Deep}.
The MMD is defined as a distance in the reproducing kernel Hilbert space (RKHS). In practice, its 
empirical version is used, as
it can be written as $Tr(\bfK \bfN)$, where
\ben
\bfK=\big[\ba{ll} \bfK^{s,s} & \bfK^{s,t} \\ \bfK^{t,s} & \bfK^{t,t}  \ea \big]  \,\, {\rm and \quad } 
\bfN=\big[\ba{rr}  \frac{1}{N^2_s}  \bfone^{s,s} & \frac{1}{N_s N_t}  \bfone^{s,t}  \\  \frac{1}{N_s N_t}  \bfone^{s,t}  
&\frac{1}{N^2_t}  \bfone^{t,t}  \ea \big],
\een
where $\bfK^{a,b}$ is the kernel distance matrix between all elements of $\bfX^a$ and $\bfX^b$,  
$\bfone^{a,b}$ is a constant matrix of size $N_a\times N_b$ with all elements being equal to 1, and
$N_s, N_t$ are the number of source and target examples. 

We integrate this loss in the total one $\Loss$ by considering the MMD between the source and target
data after the denoising. To be able to marginalize out the loss and to
keep our solution linearly solvable, we use the MMD with the 
linear kernel.  Intuitively, this corresponds to minimizing the distance between the two
centroids of the source and target data after denoising.
The corresponding loss can be expressed as follows
\begin{equation}
\Loss_{m} \equiv \frac{1}{M}  \sum_{m=1}^M Tr (\bfW^\top \tbfX_m^\top \bfN \tbfX_m \bfW).
\label{eq:3m}
\end{equation} 
After marginalizing the expected loss, we obtain $\E \Loss_3= Tr(\bfW^\top \bfM \bfW)$,  where 
$\bfM =\Exp[\tbfX^\top \bfN \tbfX]$
(see the derivations in the Appendix).

\subsubsection{\bf Reducing the MMD between the domain class means}
 
The MMD requires no labels and can be computed between all available source all target instances. 
If we have labeled source and labeled target examples
we can go one step further and modify the MMD to measure the distance between the means 
(centroids) of corresponding classes in the source and target domains~\cite{LongICCV13Transfer}. 
The corresponding loss is the following
\begin{equation}
\Loss_{c} \equiv \frac{1}{M}  \sum_{m=1}^M Tr (\bfW^\top \tbfX_m^\top \bfC \tbfX_m \bfW),
\label{eq:cm}
\end{equation}
where:
\ben
\bfC_{ij}= \left\{
\ba{ll}
     \frac{1}{N^c_s N^c_s} & {\rm if} \quad \bfx_i,\bfx_j \in \calD^s \quad \& \quad  y_i=y_j=c\\
     \frac{1}{N^c_t N^t_t} & {\rm if} \quad \bfx_i,\bfx_j \in \calD^t \quad \& \quad  y_i=y_j=c\\
     \frac{-1}{N^c_s N^c_t} & {\rm if} \quad \bfx_i\in \calD^s, \bfx_j \in \calD^t \quad \& \quad  y_i=y_j=c\\
     \frac{-1}{N^c_t N^c_s} & {\rm if}  \quad \bfx_i\in \calD^t, \bfx_j \in \calD^s \quad \& \quad  y_i=y_j=c\\
     0 & {\rm otherwise},	\ea
     \right.
\een
$N^c_s$ is the number of source instances from the class $c$ and 
$N^c_t$ is the number of target instances from the class $c$.  Note that the "otherwise" item 
above includes all cases where $y_i\neq y_j$ and where either $\bfx_i$ or $\bfx_j$ is unlabeled. 

Similarly to $\Loss_m$ in~\eq{eq:3m}, we can marginalize out the expected loss $\Loss_c$ and
obtain $\E \Loss_c = Tr(\bfW^\top \bfM_c \bfW)$, where $\bfM_c=\Exp[\tbfX^\top \bfC \tbfX]$.

\subsubsection{\bf Learning a domain classifier}
\label{sssec:domreg}

As the last option of the domain regularization $\Loss_3$ in \eqref{eq:3terms}, we explore a loss based on 
the domain classifier~\cite{ClinchantACL16Domain}. Inspired by~\cite{GaninICML15Unsupervised} who proposed 
to regularize intermediate layers in a deep learning model with a domain prediction 
task,~\cite{ClinchantACL16Domain} combines the domain prediction
regularization with the MDA. We develop a similar regularization term
for our extended framework, 
and 
use it jointly with the feature denoising term $\Loss_{1}$ and the class regularization $\Loss_{2}$.

The idea of this domain regularization is to denoise data in such a way
that pushes source data towards the target and hence allows the
cross-domain classifier to perform better on the target.  This is done
by first learning a domain classifier $\bfZ_\calD \in \Re^N$ using
a regularized ridge classifier learned on the uncorrupted data.
The regularized loss is defined as $\| \bfY_{\calD} -\bfX \bfZ_\calD\|^2 + \alpha \|\bfZ_\calD\|^2$, 
where $\bfY_{\calD} \in \Re^N$ are the domain labels (-1 for source and +1 for target).
The closed form solution is the following
\begin{equation}
\bfZ_\calD = (\bfX^\top \bfX + \alpha \eye_d)^{-1}( \bfX^\top \bfY_{\calD}).
\label{eq:Zd}
\end{equation}
Then the loss we consider in our unified framework is:
\begin{equation}
\Loss_d \equiv \frac{1}{M} \sum_{m=1}^M  \|\bfY_\calT - {\tilde \bfX_m} \bfW \bfZ_\calD\|^2,
\label{eq:d}
\end{equation}
where $\bfY_\calT=\bfone^N$ is a vector containing only ones (all
denoised instances should be predicted as target).

\section{Minimizing the total loss}
\label{sec:joint}

In the previous section we described three terms of the loss function $\Loss$.
Now we discuss two main cases of minimizing the total loss. 
First, we discuss the sequential case, where we first learn $\bfW$ using only the data 
without labels ($\Loss_1$ or $\Loss_1+\gamma \Loss_3$), and then we learn
the classifier $\bfZ_l$ or any other classifier. 
Second, we describe the joint case where $\bfW$ and $\bfZ_l$ are learned jointly, by iteratively
minimizing the total loss $\Loss_1+\lambda \Loss_2+\gamma \Loss_3$.
In both cases we consider three options for the domain regularization $\Loss_3$ and 
discuss 
three domain adaptation scenarios, {\it US, SUP} and {\it SS}.  

All mentioned combinations of the losses form different models; we denote them as follows. 
The sequential methods are prefixed by a character {\bf S} followed by the indexes
of the losses used.  For example, when we learn $\bfW$ with
$\Loss_1+\gamma \Loss_d$, the method is denoted {\bf S1D}.  When we
learn $\bfW$ and $\bfZ_l$ jointly, we prefix the method by a character {\bf J} followed by the 
loss indexes. For example, the method {\bf J12C} means
that we optimize $\Loss_1+\lambda \Loss_2+\gamma \Loss_c$.  
All the combinations are summarized in Table~\ref{tab:losses}.

\subsection{Sequential framework}
\label{ssec:sequential}


In this case, we first obtain $\bfW$ in an unsupervised manner 
and then learn a classifier $\bfZ_l$ using the 
denoised features. To get $\bfW$, we set $\lambda=0$ and minimize
$\Loss= \Loss_1 +\gamma \Loss_3$.  For each option of loss $\Loss_3$, 
we get closed-form solutions for $\bfW$, denoted {\bf S1}, {\bf S1M}, and {\bf S1D}. 
All the solutions are presented in Table~\ref{tab:losses}.
Any model can be deployed in three domain adaptation scenarios. 
In the {\it SUP} and {\it SS} cases, we additionally exploit the class labels using 
$\Loss_3 = \Loss_c$ (see {\bf S1C} in Table~\ref{tab:losses}). 

Once $\bfW$ is computed, we learn $\bfZ_l$ using \eq{eq:Zl} or use any other classifier by feeding it 
with the denoised features $\bfX_l \bfW$. In the {\it US} case, the classifiers are learned with the 
denoised source features, while in the {\it SUP} and {\it SS} cases the classifier exploits 
additionally the labeled target data.
 
\subsection{Joint framework}
\label{ssec:joint}

In this case, $\bfW$ and $\bfZ_l$ are learned {\em jointly}, by
alternatively optimizing the total loss $\Loss$ in variables $\bfW$ and
$\bfZ_l$. We start by initializing $\bfW$ with $\eye_d$ and minimize the loss
in $\bfZ_l$, then we fix $\bfZ_l$ and compute $\bfW$, and so on.
The process is repeated until convergence for a certain threshold.

The partial derivatives of $\Loss$ with respect to $\bfZ_l$ depend on $\Loss_2$ only, 
this makes solution \eq{eq:Zl} always valid. The partial derivatives with respect to $\bfW$ 
can be written as a Sylvester linear matrix equation
$\bfA \bfW + \bfW \bfB =\bfC$, that we solve using the Bartels-Stewart
algorithm~\cite{Sorensen03Direct}. Depending on which loss is used as $\Loss_3$, we
obtain three versions of the Sylvester equation, denoted {\bf J12}, {\bf J12M} and {\bf J12D} and
detailed in Table~\ref{tab:losses}.

Note that for {\bf J12D} we do not use the regularizer term
$\omega \|\bfW \|^2$ in the loss, in order to be able to reduce the partial 
derivatives to solving a Sylvester equation. 
Furthermore, in the {\it SUP}
case, as $\bfQ=\bfQ_l$ and $\bfP=\bfP_l$, if we remove
$\omega \|\bfW \|^2$ from {\bf J12} we obtain a closed form
solution $\bfW=\bfQ_l^{-1} (\bfP_l + \lambda
(1-p) \bfX_l^\top \bfY_l \bfZ_l^\top)(\eye_d+\lambda \bfZ_l \bfZ_l^\top)$.
Note that in our  experiments we found that the results with the
term (by solving a Sylvester equation) and without
(a closed form solution) are similar, but the latter case is much faster.

\section{Experimental Results}
\label{sec:eval}

In the experimental section, we pursue a number of important goals. 
First, we want to assess all models proposed\footnote{The code for all models is available at \url{http://github.com/sclincha/xrce_msda_da_regularization}}
 in the previous sections in three domain adaptation
scenarios.  Second, we evaluate the impact of the domain regularization $\Loss_3$ and the
class regularization $\Loss_2$ on the denoising matrix $\bfW$ and target classifier $\bfZ_l$.
Finally, we report the performance of the sequential and 
joint learning cases, we analyze our results and compare them to the state-of-the art.

This section is organized as follows. In Section~\ref{sec:datasets} we briefly describe three datasets used in the experiments, 
then Section~\ref{sec:settings} describes the experimental setting, including the methods and parameters used. In Section~\ref{sec:results} 
we compare the sequential and joint models with 
different loss combinations in the {\it US}, {\it SUP} and {\it SS} settings, for all datasets. 
Finally, in Section~\ref{sec:SOAres} we compare our best performing models with the state-of-the art.

\subsection{Datasets}
\label{sec:datasets}
All experiments are conducted on three domain adaptation datasets, well known in image processing 
and sentiment analysis communities.

{\bf OFF31 and OC10.} Two most popular datasets used to compare visual domain adaptation methods 
are the Office31 dataset~\cite{SaenkoECCV10Adapting} ({\bf OFF31})
and the Office+Caltech10~\cite{GongCVPR12Geodesic} ({\bf OC10}).
The former consists of three domains: Amazon ($A$), dslr ($D$) and Webcam ($W$) with images of 31 products (classes).  
The latter contains 10 of the 31 classes for the same domains and includes an extra domain from the Caltech collection. 
For all images, we use the Decaf TF6 features~\cite{DonahueX13Decaf} with the {\it full training} 
protocol~\cite{GongICML13Connecting} where all source data is used for training. 

{\bf AMT.} 
A standard dataset for textual domain adaptation is 
the Amazon dataset of text products reviews; it includes four domains: Books ($b$), DVD ($d$), Kitchen ($k$) and Electronics ($b$)
preprocessed by Blitzer \etal~\cite{BlitzerAISTATS11Domain}. Reviews are considered as positive if they have more than 3 stars, 
and negative otherwise. We adopt the experimental setting of~\cite{GaninX15Domainadversarial} where 
documents are represented by a bag of uni-grams and bi-grams with the 5000 most frequent common words selected 
and a tf-idf weighting scheme. 

\subsection{Methods and settings}
\label{sec:settings}

Most models proposed in the previous sections produce the denoising matrix $\bfW$ and the target classifier $\bfZ_l$, which can be inferred sequentially or jointly. The sequential approach learns first the 
matrix $\bfW$ and then a target classifier $\bfZ_l$ for a fixed $\bfW$. The joint approach learns $\bfW$ and $\bfZ_l$ 
jointly\footnote{In this case, while we do not have the guarantee a global minimum,  we observed  in general to quick convergence of the loss (only a few iterations).} 
by alternating the updates of $\bfW$ and $\bfZ_l$ using the same loss $\Loss$.  

For each dataset, we consider all domains and take all possible source-target pairs as domain adaptation tasks. 
For example, for {\bf OFF31} with three domains, we consider six following adaptation tasks: 
 D$\rar$A, W$\rar$A,  A$\rar$D, W$\rar$D, A$\rar$W, and D$\rar$W. Similarly, for {\bf OC10} and {\bf AMT} which include four domains each, we consider 12 different source-target pairs as adaptation tasks. To compare the different models, we report \textbf{averaged} accuracies over all adaptation tasks for a given dataset.  
In the supervised ({\it SUP}) and semi-supervised ({\it SS}) scenarios,
we randomly select 3 target instances per class to form the target training set, and use the rest for the test. 
 
In addition to the sequential and joint model learning, we include in our framework two standard classifiers, 
the nearest neighbor (NN) classifier and the Domain Specific Class Means (DSCM) as 
they represent a valuable alternative to the ridge target $\bfZ_l$ classifier. 
classifier~\cite{CsurkaTASKCV14Domain}. In DSCM, a target test example is assigned to a class by using a soft-max 
distance to the domain specific  class means. The main reason for choosing these classifiers is that NN is related to retrieval 
	(equivalent to precision at \@1) and NCM with clustering, so the impact of $\bfW$ on these two extra 
	tasks is indirectly assessed. 
	
Fed with the denoised instances, obtained with matrix $\bfW$, these classifiers help assess the value of our 
framework for domain adaptation tasks. 

To ensure the fair comparison of all methods, we run all experiments with a unique parameter setting. All selected parameter values are explained below: 
Besides, cross-validation on the source is not the best way to set model parameters for transfer learning and domain adaptation \cite{ZhongECML10Cross}.

\bi
\item we set $\lambda=\gamma=1$ as term weights, this corresponds to the equal weighting in the global loss \eq{eq:3terms}; 
\item we set $\omega=10^{-2}$ as $\bfW$ norm regularization in \eq{eq:WMDA} (as in~\cite{ChenICML12Marginalized}); 
\item we set the dropping noise level for $\bfP$ and $\bfQ$ in \eq{eq:WMDA}: $p=0.5$ for image datasets and $p=0.9$ for AMT, as text representations are initially very sparse and a higher noise level is required;
\item we set $\alpha=\delta=1$ for the classifier regularization terms in~\eq{eq:mcf} and \eq{eq:Zd};
\item we consider a single layer MDA only, to enable a fair comparison of different loss combinations and learning 
methods\footnote{Similarly to the stacked MDA, our framework can be extended to a stacked version, 
	where the denoised features $\bfX \bfW$ of one layer become the input of a next layer
	and nonlinear functions such as tangent hyperbolic are applied between the two.}.
\ei

To reveal all strong and weak points of our framework, we compare all models and classifiers with two baselines. 
The first baseline is denoted {\bf BL} and provided by the classifier learned on the original features, without denoising. 
The classifier $\bfZ_l$ is learned using \eq{eq:Zl} with $p=0$ and hence $\bfQ=\bfS$. 
The second baseline refers to the original MDA method~\cite{ChenICML12Marginalized} and corresponds to {\bf S1} method in our framework. 
It uses the loss $\Loss_1$ to build $\bfW$ in an unsupervised manner and learns a classifier on the denoised features. 
 
In the following subsections, we compare the methods of our framework to the baselines, for all domain adaptation 
settings and all datasets. 
 
\subsection{Comparing domain adaptation methods}
\label{sec:results}

\subsubsection{\bf Unsupervised Setting}
\label{sssec:USres}

In this case, labeled data are available from the source domain only, $\bfX_l=\bfX^s$. 
We compare the different models described in Section~\ref{sec:joint} (see the summary in Table~\ref{tab:losses}). 
For each method and each domain adaptation task, we learn the $\bfZ_l$ classifier and the NN and DSCM 
classifiers applied to the denoised features. The accuracy results are averaged per dataset and reported in Table~\ref{tab:unsupervised_adaptation}. 
 
The first observation is that all {\bf BL} baselines get improved by the MDA ({\bf S1}).
Second, on the text data ({\rm AMT}), the $\bfZ_l$ classifier performs the best\footnote{NN and DSCM results are not included in the table, due to space limitation.}, Third, 
on the image collections the picture is more complex, 
with the NN showing the globally highest accuracy.
If we compare the baseline {\bf S1} and extended methods, we can conclude the following. In {\it sequential framework}, the domain regularization {\bf S1D} often improves over {\bf S1} for the linear classifier $\bfZ_l$ and the DSCM, but not of the NN. In the {\it joint framework}, the class regularization $\Loss_2$ degrades the linear classifier $\bfZ_l$ results but improves the DSCM classifier.

To conclude, in the unsupervised domain adaptation, the best strategy may depend on the data type, regularization and classification method. If we compare methods by averaging their results 
over the rows in Table~\ref{tab:unsupervised_adaptation}, {\bf S1D} with the average 75.7\% appears as the best 
strategy over all classifiers and datasets, followed by {\bf J12}. Both outperform the baseline {\bf S1} with the average 74.5\%. This suggests that the best strategy is to learn the denoising $\bfW$ with
the domain regularization and then to learn any classifier from denoised features.

 \begin{table}[ttt]
 \footnotesize
 \begin{center}
 \caption{Unsupervised domain adaptation. 
 Bold indicates best result per dataset, underlined are improvements over \bf{S1}.}
\label{tab:unsupervised_adaptation}
\begin{tabular}{|l|c c c|c c c|c|} 
  \hline 
  & \multicolumn{3}{|c}{\bf OC10} & \multicolumn{3}{|c|}{\bf OFF31} & {\bf AMT} \\
 \hline 
     &  \small{nn}   & \small{dscm}      &   \small{$\bfZ_l$}       &  \small{nn}  &    \small{dscm}  &   \small{$\bfZ_l$}  &  \small{$\bfZ_l$}   \\
 \hline
 \bf{BL}       &  84.5 &  78.7     & 82.6                   &  65.2 &   61.6 &   62.8             &  76.7   \\
 \bf{S1}       &  85.2 &  79.9     & 84                     &   65.9 &  62.1 &   66.5             & 81.6    \\
 \hline
 \bf{S1D}    &  84.9   & \ul{80.9}  & \ul{\textbf {85.6}}         & 65.8  & \ul{62.6}&   \ul{67.1}     & \textbf{\ul{82.2}}    \\
 \hline
         &           &         &          &           &         &         \\
  \bf{J12}     & 84.4 &  \ul{82.3}     &  82.3          &   \ul{\textbf{67.7}}  &  \ul{65} &  65       &  76.0   \\
  \hline
  \bf{J12D}    & 82.8  &  \ul{83.2}     &  83.4           &   64.4    &   \ul{65.2}   &    65.5     &  76.0   \\ 
   \hline 
\end{tabular}

\end{center}
\end{table}

\subsubsection{\bf Supervised Setting}
\label{sssec:Supres}
 
In this case we have $\bfX=\bfX_l=[\bfX^s,\bfX_l^t]$, $\bfQ_l=\bfQ$ and unlabeled target data is unavailable.
In Table~\ref{tab:supervised_adaptation} we report the evaluation results for the different models using 
$\bfZ_l$, NN and DSCM classifiers. It is easy to note that all models in the supervised case behave quite different from 
the unsupervised one.

{\it Sequential framework.} The domain regularization $\Loss_d$ seems to harm the baseline performance. 
One possible explanation is that only few target examples are available, comparing to the much larger source set. 
In contrast, adding the regularization $\Loss_c$ that exploits the class labels improves in many cases 
the results.

{\it Joint framework.} Adding $\Loss_2$ improves  the results in general, except when using $\bfZ_l$ on AMT dataset. 
While {\bf J12} and {\bf J1D}  perform rather similarly, {\bf J12C} outperforms them globally on all datasets and classifiers. 
Note however, that the jointly learned $\bfZ_l$  performs less well than using the jointly learned 
$\bfW$ to denoise the data on which a standard classifier is learned. This is the case when
the framework implicitly combines the benefit of the joint learning with any classifier; 
such a strategy seems to work the best. 

\begin{table}[ttt]
 \begin{center}
 \footnotesize 
  \caption{Supervised domain adaptation. 
 Bold indicates best result per dataset, underlined are improvements over \bf{S1}.}
\label{tab:supervised_adaptation}
\begin{tabular}{|l|c c c|c c c| c|} 
  \hline 
  & \multicolumn{3}{c}{\small \bf OC10} & \multicolumn{3}{|c|}{ \small{\bf OFF31}} & { \footnotesize{\bf AMT}} \\
 \hline 
    &  \small{nn}   & \small{dscm}      &   \small{$\bfZ_l$}       &  \small{nn}  &    \small{dscm}  &   \small{$\bfZ_l$}     &   \small{$\bfZ_l$}   \\
 \hline
 \textbf{BL}             &   90.8       &  91.6       & 88.1           & 77.6             & 76.6     & 70.3           & 77.3 \\
 \textbf{S1}     &   90.6       &  \bf{92.1}  & 89.7           & 77.2             & 76.2     & 72.1            & \bf{78.4} \\
 \hline 
 \textbf{S1D}         &  89.8      &  91.6       & 89             & 76.7             & 75.9     & 71.9             & 74.3 \\
 \textbf{S1C}     &   \ul{90.7} & 92          & \ul{90.1}      & 76.8             & \ul{77}  & \ul{72.7}    & \bf{78.4}  \\
 \hline 
 \textbf{J12}     &\ul{90.7}       & 92          &  \ul{90.1}   & \ul{79.1}       & \ul{80.7} & \ul{78.0}   & 71.1    \\
 \hline
 \textbf{J12D}     &\ul{90.7}      & 90.1           &\ul{89.9}  &  \ul{79.1}      & \ul{80.6} & \ul{78.0}   & 71.1    \\
 \textbf{J12C}     &\bf{\ul{92.1}}  &   \bf{92.1}   &\ul{91.1}  & \bf{\ul{80.8}}  & \ul{80.7} & \ul{78.1}   & 73.9     \\
  \hline 
 \end{tabular}
 
 \end{center}
 \end{table}

\subsubsection{\bf Semi-supervised Setting}
\label{sssec:SSres}

In this case, a large set of unlabeled target data is available together with a small set of labeled target data.
We explore if the proposed methods are able to take advantage of the two. 
Table~\ref {tab:semi_adaptation} shows the results of different models on the three datasets, and  them to the baselines.

{\it Sequential framework.} Compared to the supervised case, having more target 
 examples makes the domain regularization $\Loss_3$ to either have no effect or slightly 
 improve the results. 

{\it Joint framework.} Adding class regularization $\Loss_2$ either improves or does not change the 
results, except learning $\bfZ_l$ for AMT, where a significant drop is observed. Adding $\Loss_d$ often decreases the performance, while adding $\Loss_c$ is less harmful. 
In general, in the semi-supervised case {\bf J12} is the best strategy for the image datasets. For the text dataset, like in the  unsupervised case, using {\bf S1D} with $\bfW$, followed by learning a classifier on the denoised features seems to be a better option.

\begin{table}[ttt]
\footnotesize
 \begin{center}
 \caption{Semi-supervised domain adaptation on OC10, OFF31 and AMT. 
Bold indicates the best result per dataset, underline indicates the improvement over \bf{L1}.}
\label{tab:semi_adaptation}
 \begin{tabular}{|l|c c c|c c c|c|} 
  \hline 
  & \multicolumn{3}{c}{\bf OC10} & \multicolumn{3}{|c|}{\bf OFF31} & {\bf AMT} \\
 \hline 
     &  \small{nn}   & \small{dscm}      &   \small{$\bfZ_l$}       &  \small{nn}  &    \small{dscm}  &   \small{$\bfZ_l$}     &  \small{$\bfZ_l$}   \\
 \hline
 {BL}            &   90.8        &  91.6     & 88.1    &  77.6          & 76.6         & 70.3           & 77.3    \\
 \textbf{S1}     &   91.1        &  91.9     &  90.4   &  78.4          & 76.9         & 74.6           & 82.4   \\
 \hline 
 \textbf{S1D}     &  \ul{91.2}   &  91.9     &  \ul{90.7}   &  78.1          & \ul{77.3}     &  74.5         &  \bf{\ul{82.7}}  \\
 \textbf{S1C}     &  \ul{91.3}   &  \ul{92.1}     &  90.4   &  \ul{78.5}     & \ul{77.6}     &  74.5         &  82.4  \\
 \hline 
 \textbf{J12}     &  \textbf{\ul{92.3}}   &  91.8     &   89.1  & \textbf{\ul{80.8}} & \ul{80.5}    &  \ul{74.7}     &   76.6 \\
 \hline
 \textbf{J12D}     & 87.9        &   89.8    &   89.1   &  \textbf{\ul{80.8}}    &  \ul{80.4}    &  \ul{74.7}         &   76.6  \\
 \textbf{J12C}     & \ul{92.1}   &   91.6    &   89.1   &   \textbf{\ul{80.8}}   &  \ul{80.2}   &  \ul{74.9}          &     76.6 \\
  \hline 
 \end{tabular}
 \end{center}
 
 \end{table}

\subsection{Comparison with the state of the art}
\label{sec:SOAres}

We complete the experimental section by comparing our results to the state of art results,
in the unsupervised and semi-supervised cases\footnote{We exclude the supervised scenario as rarely addressed in the literature.}.

\noindent{\bf OC10.} Using the FC6 features, our {\bf S1M} accuracy (86.5\%) in Table~\ref{tab:unsupervised_adaptation} 
 outperforms\footnote{We report results from~\cite{FarajidavarACCV14Transductive}.} the domain adaptive SVM~\cite{BruzzonePAMI10Domain} (70.3\%) and 
 domain adaptation via auxiliary classifiers~\cite{DuanICML09Domain} (84\%), and slightly underperforms the more complex 
 JDA~\cite{LongICCV13Transfer} (87.5\%) and TTM~\cite{FarajidavarACCV14Transductive} (87.5\%) methods. \\

\noindent {\bf OFF31.} Table \ref{tab:OFFSOA} compares our {\bf S1D}+$\bfZ_l$  and 
{\bf S12}+NN results with the feature transformation methods, using the same deep FC6 
features, and with recent deep learning methods. It reports the results for six domain adaptation tasks 
on {\bf OFF31} and their average.
We can see that our methods behave similarly or better than feature transformation methods, but below the
deep adaptation ones. Designed to be fast, our methods solve a few linear systems at training time and 
a simple matrix multiplication at test time, while the deep architectures have thousands of parameters to be tuned with the back propagation and GPU computations at the training time. \\

\noindent {\bf AMT.} Our {\bf L1D}+$\bfZ_l$ results (82.2\%, see Table~\ref{tab:unsupervised_adaptation})
is similar to the state-of-the art results obtained with the Domain-Adversarial Neural Networks (DANN)~\cite{GaninX15Domainadversarial}, despite  the fact that DANN uses a 5 layer stacked MDA where the 5 
outputs are concatenated with input to generate 30,000 dimensional features, on which the network is trained. 
\begin{table*}[ht]\begin{center}
\footnotesize
\caption{Unsupervised adaptation on the {\bf OFF31}: (top) feature transformation methods with FC6 features; 
(middle) our results with using  FC6 features; (bottom) the deep domain adaptation methods. Best results per task are in bold. Underline indicates best results on FC6 features.}
\label{tab:OFFSOA}
\bt{|c|ccccccc|}
\hline
 & A$\rar$W & D$\rar$W & W$\rar$D  &  D$\rar$A & W$\rar$A & A$\rar$D & Mean \\
 \hline
GFK+SVM~\cite{SunAAAI16Return} & 37.8 & 81.0 & 86.9  & 34.8 & 31.4 & 44.8  & 49.1  \\
SA+SVM~\cite{SunAAAI16Return} &   35 &  74.5 & 81.5 & 32.3 &  30.1 & 41.3 & 49.1\\
TCA+SVM~\cite{SunAAAI16Return} & 36.8 & 82.3 & 84.1 & 32.9 & 28.9 & 40.6  & 50.9 \\
CORAL+SVM~\cite{SunAAAI16Return} & 48.4 & \ul{96.5} & \ul{\bf{99.2}}  & 44.4 & 41.9 & 53.7 & 64.0 \\
\hline
{\bf L12} + NN & 50.5 & 94.3 &  98.0 & 46.4 & 42.8 & 53.8 & 64.3 \\
{\bf L1D} + RDG & \ul{53.9}  &  91.2 & 95.1  & \ul{47.3} & \bf{49.0} & \ul{55.5} & \ul{65.3} \\
\hline
 DDC~\cite{TzengX14Deep} & 59.4 & 92.5 & 91.7 & 52.1 & 52.2 & 64.4  & 68.7  \\
RGrad~\cite{GaninX15Domainadversarial} & 67.3 & 94.0 & 93.7 & - &- & - &-\\
DAN~\cite{LongICML15Learning} & \bf{68.5} & \bf{96.0} & 99 & {\bf 54}& \bf{53.1} & \bf{67.0} & {\bf 72.9} \\
 \hline
\et
\end{center}
\end{table*}

Concerning the semi-supervised scenario, it is much less used and most papers report results with SURF BOV features and the 
sampling protocol ~\cite{SaenkoECCV10Adapting,GongCVPR12Geodesic}. We therefore tested our methods on {\bf OC10}
with {\bf L12C}+DSCM and BOV features averaged over the 20 random samples; and we get an accuracy of 55.8\% that is above most 
state of art results, including GFK~\cite{GongCVPR12Geodesic} (48.6\%), SA~\cite{FernandoICCV13Unsupervised} (53.6\%), MMDT~\cite{HoffmanICLR13Efficient} (52.5\%).

\section{Conclusion}
\label{sec:conclusion}

We proposed an extended framework for domain adaptation, where the
state-of-the-art marginalized denoising autoencoder is extended with
domain and class regularization terms, aimed at addressing
unsupervised, supervised and semi-supervised scenarios.  The domain
regularization drives the denoising of both source and target data
toward domain invariant features. Two families of domain
regularization, based on domain prediction and the
maximum mean discrepancy, are proposed.  The class
regularization learns a cross-domain classifier jointly with the
common representation learning.  In all cases, the models can be
reduced to solving a linear matrix equation or its Sylvester version,
for which efficient algorithms exist. 

We presented the results of an extensive set of experiments on two image and one 
text benchmark datasets, where the proposed framework is tested in different
settings.  We showed that adding the new regularization terms allow to
outperform the baselines and help design best performing
strategies for each adaptation scenarios and data types. Compared to the
state of art we showed that despite of their speed and
relatively low cost, our models yield comparable or better results
than existing feature transformation methods but below highly expensive
non-linear methods with additional data processing such as the landmark
selection or those using deep architectures requiring costly
operations both at training and at test time.
Furthermore, similarly to the stacked MDA framework, we can
easily stack several layers together with only forward learning, where
the denoised features of the previous layer become the input of a new
layer and nonlinear functions can be applied between the layers.


\section*{Appendix}

In this section we derive and show the partial derivatives of each expected loss terms according to $\bfW$ and 
when relevant according to $\bfZ_l$. In our derivations we used the fact that the trace is linear
and it commutes with the expectations and used the derivative formulas of the trace from \cite{PetersenB12Matrix}: 
\ben
\Loss_1  & =& \Exp[  Tr((\bfX -\tbfX \bfW)^\top (\bfX -\tbfX \bfW))] + \omega \|\bfW \|^2    \\
& = &  Tr(\bfX^\top\bfX) - 2 Tr(\Exp[ \bfX^\top\tbfX ]\bfW)  \\
& + & Tr(\bfW^\top E[\tbfX^\top \tbfX] \bfW) +\omega \|\bfW \|^2 \\
&  = & \|\bfX\|^2 - 2Tr(\bfP \bfW) + Tr (\bfW^\top \bfQ \bfW) +\omega \|\bfW \|^2
\een
where $\tbfX$ is the random variable representing the corrupted
$\bfX^m$ features, $\Exp[\tbfX]=(1-p)\bfX$, $\bfP=\Exp[\bfX^\top \tbfX]$ and $\bfQ=\Exp[\tbfX^\top \tbfX]$. 
If we denote by $\bfS$ the covariance matrix  $\bfX^{\top} \bfX$ of
 the uncorrupted data,   we have $\bfP=(1-p)\bfS$ and:
\ben
\bfQ_{ij} =\left[ 
  \ba{ll} 
    \bfS_{i j} (1-p)^2 , & {\rm if} \quad i \neq j,\\
    \bfS_{i j} (1-p),     & {\rm if} \quad i = j.
  \ea \right.
\een
The partial derivatives of $\Loss_1$ can be written as:
\ben
\frac{\partial \Loss_1}{\partial \bfW} & =  & - 2 \bfP + 2(\bfQ +\omega \eye_d)\bfW  \label{eq:DWL1}
\een

Note that $\Loss_2$ \eq{eq:mcf} and $\Loss_d$ \eq{eq:d} are similar (we have $\bfZ_{\cal{D}}$ instead of $\bfZ_l$,  $\bfY_\calT$ instead of 
$\bfY_l$  and $\delta=0$).  Therefore we derive here the expected loss and its derivatives derivatives only for $\Loss_2$:
\ben
\Loss_2 & =&   \Exp[Tr((\bfY_l - {\tilde \bfX_l} \bfW \bfZ_l)^\top (\bfY_l - {\tilde \bfX_l} \bfW \bfZ_l))]  + \delta \|\bfZ_l\|^2  \nonumber \\
& =&  Tr(\bfY_l^\top\bfY_l) -2 Tr(\bfY_l^\top \Exp[\tbfX_l]\bfW \bfZ_l)  \nonumber \\
& + &  Tr(\bfZ_l^\top \bfW^\top \Exp[\tbfX_l^\top \tbfX_l] \bfW \bfZ_l) + \delta \|\bfZ_l\|^2  \nonumber \\
&=& \|\bfY\|^2 -2 (1-p) \bfY_l^\top \bfX_l \bfW \bfZ_l \\
&+& \bfZ_l \bfW^\top \bfQ_l \bfW \bfZ_l+ \delta \|\bfZ_l\|^2 
\een
where  $\bfX_l$ is the labeled part of the data $[\bfX^s,\bfX^t_l]$, where $\bfX^t_l$ is empty 
 in the unsupervised scenario and  $\bfQ_l$ is computed as $\bfQ$ but with $\bfS_l=\bfX^{\top} \bfX$.
In the case of $\Loss_2$ (but not $\Loss_d$) we derive the partial derivatives also according to $\bfZ_l$:
\ben
\frac{\partial \Loss_2}{\partial \bfW} & =  & -2 (1-p) \bfX_l^\top \bfY_l \bfZ_l+ 2 \bfQ_l \bfW \bfZ_l \bfZ_l^\top  \\
\frac{\partial \Loss_2}{\partial \bfZ_l} &= & -2 (1-p) \bfW^T \bfX_l^\top \bfY_l+2 (\bfW^\top \bfQ_l \bfW + \delta \eye_d)\bfZ_l 
\een
 
Finally, in the case of MMD the marginalized loss becomes:
\ben
\Loss_{m} &=&Tr (\bfW^\top \Exp[\tbfX^\top \bfN \tbfX] \bfW)=Tr(\bfW^\top \bfM \bfW)
\een
yielding to the partial derivatives 
$\partial \Loss_m/\partial \bfW= 2 \bfM \bfW$, 
where $\bfM =\Exp[\tbfX^\top \bfN \tbfX]$ can be computed as $\bfQ$
 in \eq{eq:PQ} using $\bfS_m=\bfX^\top \bfN \bfX$ instead of
 $\bfS$.  For $\Loss_c$ we have the loss equal to  $2 \bfM_c \bfW$,
where  $\bfM_c$ is computed with  $\bfS_c=\bfX_l^\top \bfC \bfX_l$ .


\end{document}